\documentclass[runningheads]{llncs}
\usepackage[T1]{fontenc}
\usepackage{color}
\usepackage{graphicx}
\usepackage{subfigure}
\usepackage{mathtools}
\newcommand\numberthis{\addtocounter{equation}{1}\tag{\theequation}}
\usepackage[linesnumbered,ruled,vlined]{algorithm2e}
\SetKwInOut{Input}{Input}                
\SetKwInOut{Output}{Output}              
\SetKwInOut{Init}{Initialize}
\begin{document}
\title{A Multi-Domain Multi-Task Approach for Feature
Selection from 
Bulk RNA Datasets}
%
%
\author{Karim Salta\inst{1}\orcidID{0009-0002-4589-3335} \and
Tomojit Ghosh\inst{2}\orcidID{0000-0001-6875-3840} \and
Michael Kirby\inst{3}\orcidID{0000-0001-9802-9263}}

\authorrunning{K. Salta et al.}
%
\institute{Colorado State University, Fort Collins CO 80523, USA \\
\email{karim.karimov@colostate.edu} \and
University of Tennessee, Chattanooga TN 37403 USA Z \\
\email{tomojit-ghosh@utc.edu} \and
Colorado State University, Fort Collins CO 80523, USA \\
\email{michael.kirby@colostate.edu }}

\maketitle              
\begin{abstract}
In this paper a multi-domain multi-task algorithm for feature selection in bulk RNAseq data is proposed. Two datasets are investigated arising from mouse
host immune response to  {\it Salmonella} infection. 
Data is collected from several strains of collaborative cross mice.  Samples  from  the spleen and liver
serve as the two domains.
Several machine learning experiments are conducted and the small subset of discriminative across domains features have been extracted in each case. The algorithm proves viable and underlines the benefits of across domain feature selection by extracting new subset of discriminative features which couldn't be extracted only by one-domain approach.

 
\end{abstract}
\section{Introduction}
In the field of bioinformatics, researchers often use microarray or next-generation sequencing techniques to study the expression levels of genes, with each sample  typically having tens of thousands of features. The large number of features often
necessitates the use of feature selection algorithms to improve the performance of machine learning tasks such as classification given that many of the observed features
may be unrelated to the biological phenomenon of interest. 
In this way, feature selection algorithms can be used to determine the processes related
to a biological mechanism, e.g., the host immune response to infection.   
Other downstream benefits of feature selection include data visualization and understanding, reduced storage requirements, and faster computations. 

Multi-domain feature extraction addresses the problem of leveraging data from disparate sources and is related to the more general problem of multi-domain learning (MDL) \cite{MDL}.
In this paper we address a special case of MDL that
involves the classification of data related to the host immune response to infection.  The host consists of multiple lines of the Collaborative Cross
mouse, the pathogen under consideration is Salmonella.  The two domains consist of gene expression data collected from liver and spleen tissues.
The proposed machine learning approach has the potential 
to identify novel biomarkers whose signals are
are too weak to be captured
by analyzing domains individually.
The methodology will be demonstrated by selecting potentially important biomarkers that appear to be amplified in strength by the multi-domain data synthesis for characterizing biological processes that exist
simultaneously in different tissues.

In a variety of existing methods the designs vary from shallow to deep, with the networks optimized for regression, classification, dimensionality reduction or a combination of multiple tasks, i.e. multi-task learning  (MTL) \cite{MTL}. Most of the feature selection methods fall into the category of MLT methods, with the the objective function considering the combination of different goals  resulting in better generalization of results. However, the majority of them focus on a single domain feature selection.
In this this paper we suggest a new MDL method with a multi-task objective (MDL/MTL) function, or, in terms of \cite{MDMT}, a multi-domain multi-task (MDMT) method. The MDMT methods have been used widely in Natural Language Processing applications but much less for the analysis of biological data sets.

The feature selection task  is frequently performed by the introduction of $l_p$-norms with $p=0, 1, 2$, or a  combination of norms of weights at intermediate layers, possibly stacked closer to the input. 
The $l_0$ is of course appealing but the most expensive to compute~\cite{DFS,SINNHNRC}.
However, in \cite{SCE} it was shown that $l_1$-norm based methods can be quite competitive when applied to biological data, while not having the complications of dealing with $l_0$-norm. When applied to biological task, the $l_0$-norm based methods may require {\it a priori} knowledge of the size of a subset of biologically significant features, which is certainly not the case for many explorative tasks. Moreover, when the domains are very different the subsets of important features can be very different as well, hence it becomes even harder to approximate the number of selected features. At the same time, $l_0$-norm based methods haven't been battle-tested across domain as much as $l_1$-norm based methods, intuitively, the discrete distribution can negatively effect the alignment of disparate domains.  Considering all the pros and cons, in this paper we opt to employ $\ell_1$-norm based sparsity promotion.

The contributions of this paper include the following:  we propose a new
sparsity promoting MDMT architecture for feature selection;  this approach uses a new masking
term that restricts the features that
contribute to the cost function; 
we demonstrate the utility of developed algorithm on gene expression  dataset including liver and spleen domains; we conclude that our MDMT approach allows to find new features that are significantly discriminative only across two domains, i.e., are identified when the data is 
restricted to a single domain. The promise 
of this approach is an enriched picture of
the host immune response that has the potential
to lead to a better understanding of the biological process {\it across tissues}.

The organization of this paper is as follows:
In \textbf{Related Work}, we present a review of the related articles, situating our study within the broader context of the field and highlighting key contributions from prior research.  \textbf{Methodology} details the method we employed, outlining the design of a neural network behind our chosen approach and the techniques utilized for design. In \textbf{Data}, we describe the data used in our investigation, elaborating on pre-processing steps and labelled groups. \textbf{Experiment} delves into the computational experimental setup, discussing the training process, and criteria for rough-tuning. In \textbf{Results}, we present the results of our experiments, providing an interpretation and discussion of our findings. Finally, in \textbf{Conclusions}, we summarize the highlights of the research
and contributions, the implications of our results, and potential avenues for future work in this domain.

\section{Related Work} 

The majority of the feature selection methods study one domain. In what follows we survey a variety of different techniques and algorithms, including linear, non-linear methods and the methods exploiting neural networks. In one of the most influential papers \cite{Lasso},  feature selection is cast as a regression task with $l_1$ regularization of the norm of discriminative vector. This has become a common approach, see also, \cite{AdaptiveLasso,RelaxedLasso,FusedLasso,ElasticNet}. The Lasso-type methods fail to capture nonlinear interactions between the features. The non-linear methods developed as the kernelized modifications of Lasso method showed decent efficiency when applied to biological data \cite{HSICLasso,NonParKer,LLR}. 
Note that there are also some other methods aiming to sparsify a signal in latent dimensions \cite{Latent1,Latent2}.

Deep Feature Selection
DFS \cite{DFS} is one of the first deep neural network algorithms
designed specifically for feature selection.  DFS employs a one-to-one sparsity layer at the input.
The weights on these single connections are penalized with a $\ell_1$-norm minimization 
of norm of weights of this layer in a spirit of \cite{ElasticNet}; the resulting non-zero weights
in the sparsity layer correspond to the selected features.
In \cite{AEUFS}  autoencoders are proposed for feature selection with a  sparsity layer 
used in a fashion similar to DFS.
Now the $\ell_1$-norm of the weights of sparsity layer is minimized jointly with the reconstruction error. Concrete Autoencoders (CAE) \cite{CAE} use a concrete selector layer as the first layer in autoencoder setting based on continuous relaxations of concrete random variables suggested in \cite{CAEbase}. 
A supervised CAE method reported in \cite{FsNet} was apparently susceptible to overfitting with limited data.
The FsNet paper \cite{FsNet} addressed this problem by introducing small weight-predictor networks. In terms of design \cite{FsNet} is one of the closest to the design developed in this paper, i.e., our method is based on two neural networks: autoencoder and classifier in latent space, but the approach to sparsification is different, and, most importantly, the method in \cite{FsNet} is designed for one domain and the implication of extracted features is quite different.

Most of the feature selection methods can be grouped into three broad categories: filter, wrapper, and embedded methods. In filter methods the features are typically scored, ranked and thresholded with respect to some classification task using different measures such as correlation and mutual information \cite{filters}. Filter methods can be very fast, but the quality of extracted features is poor in terms of robustness and adaptability to different datasets. The wrapper methods \cite{wrappers} are universal methods used on top of any learning algorithm based on practical heuristic search of a subset of $d$ features in $2^d$ space providing the better performance for the underlying algorithm. They are universal and capable of obtaining great results given the large number of samples. For small datasets in high-dimensional space they tend to overfit, and the NP-hardness of the problem makes the computations prohibitively expensive. In embedded methods the feature selection process is typically performed concurrently with some learning algorithm. For example, Iterative Feature Removal (IFR) uses the absolute weights of a sparse SVM model as a criterion for selecting features from a high-dimensional biological data set \cite{Hara}. Our paper along with the most related works falls into the embedded method category.

\section{Methodology}
The methods described above only address data residing in one domain. The major question that motivated this study was what are the biological features in datasets sampled from different domains that appear to be related to the host immune response to infection only when studied across domains, naturally leading the consideration of the domain alignment task along with the feature selection. With that said and with a general design of the network in mind we've been looking for the most capable in terms of domain alignment method in application to RNA data. This led us to \cite{Uhler}, which is an MDMT method based on a pair of domain specific variational autoencoders (VAE's) \cite{VAE} generating aligned embeddings for datasets of very different modalities (singe-cell RNA and Chromatin images), and we adapt this method now enhanced with sparsity promoting optimization
constraints for feature selection. In order to find a universal representation across tasks, the MTL methods in deep neural networks \cite{MTLDeep} either improve the architecture of neural networks, or try to find a balance between concurrently trained objectives. This paper benefits from both since our network has shared subnets and at the same time we roughly fine-tune the coefficients used in \cite{Uhler} along with the contribution of sparsity promoting loss function. Utilizing both methods is also justified by the results of generalization of unbalanced optimization methods, e.g. \cite{balance1}, \cite{balance2}, \cite{balance3}, indicating that overall they don't outperform the naive approach when all loss functions are weighted with constant scalars.

\begin{figure}[!htbp]
\center
\includegraphics[width = \textwidth, height=0.3\textheight]{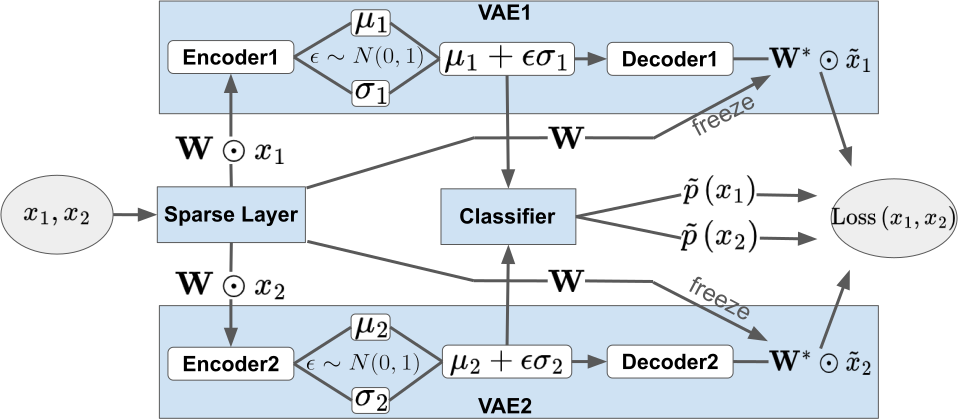}
\caption{Network design}
\label{fig:diagram}
\end{figure}

Our proposed method is based on the network shown in Figure~\ref{fig:diagram},  implemented in PyTorch and trained with the AdamW optimizer. The cost function
is a weighted combination of objective functions associated with  
the reconstruction, classification and sparsification tasks.
In our settings we can observe that during the early stages of training the algorithm is learning the shared representations of different domains such that similar samples group together regardless of domain of origin. 
At the later stages, the sparsification goal is becoming more important with a relatively higher contribution to the total objective and this behaviour continues until the conflicting tasks reach the balance and no further sparsification is possible without a significant loss in classification. We  run the training process multiple times. In a spirit of embedded methods, we treat the resulting magnitude of the weights of sparse layer as indicating the importance of different features. However, for the post-processing 
we employ the frequency of selected features across all runs as
it appears to be a more robust metric for feature importance.

It was mentioned before that our design is developed based on the network suggested in \cite{Uhler} with modifications. The classication task is performed in latent space as before, but the inputs of domain-specific VAE's are sparsified by the shared Sparsification Layer  (\textbf{SL}) and, naturally, the VAE's are trained to reconstruct only these sparsified inputs. Note that in \cite{Uhler}, 
a primary goal is the 
alignment of the domains in the latent space 
coupled with the  reconstruction of hyper-dimensional RNA data. 
In contrast to \cite{Uhler},  
our main goal is the across-domains classification with the sparsification of inputs. The choice of variational modification of autoencoders was dictated by the distribution of latent space provided by this particular modification, allowing further indirect "easy" and relaxed alignment through the shared classifier. Hence, we not only train VAE's solely to reconstruct the sparsified signal, but also we omit the loss function minimizing the KL-divergence across domains from the original design.
The resulting network consist of 4 subnets:  shared between domains Sparse Layer (\textbf{SL}) and \textbf{Classifier} subnets, and two domain specific Variational AutoEncoders (\textbf{VAE1} and \textbf{VAE2}),  as
depicted in Figure~\ref{fig:diagram}. 

The \textbf{SL} is a one-to-one mapping: $x \rightarrow \mathbf{W} \odot x$, with the $\ell_1$-norm of weights $\left\|\mathbf{W}\right\|_1$ penalty used to promote sparsity.

The \textbf{VAE1} and \textbf{VAE2} subnets are deep fully-connected networks with the following specifications:
\begin{itemize}
    \item{\textbf{Encoder}:} 2 linear layers with 1024 nodes with batch normalization and Relu-activations, followed by 1 linear mapping to $\mu$ and $\sigma$ living in 128-dimensional space
    \item{\textbf{Decoder}:} 2 linear layers with 1024 nodes with batch normalization and Relu-activations, followed by 1 layer mapping to input space
\end{itemize}
Each \textbf{Encoder} performs the mapping of the sparsified input to $\mathbf{R}^{128}$: $\mathbf{W} \odot x \rightarrow \mu, \, \sigma$, while the \textbf{Decoder} maps distribution in the latent space into the input space: $\mu + \sigma \rightarrow \tilde{x} \in \mathbf{R}^{34861}$, and the output is further masked by the frozen weights of the sparse layer $\mathbf{W}^* \odot \tilde{x}$ and fed to the MSE loss function of respective VAE.

The \textbf{Classifier} is the subnet consisting of 5 linear layers with 1024 nodes with Relu-activations, followed by 1 layer mapping to 2-dimensional space and 1 layer mapping to 1-dimensional space followed by a standard sigmoidal activation function. It is trained to classify embeddings of inputs from both domains in their respective latent spaces.

The overall objective function is a weighted sum of objective functions for different tasks including reconstruction, normalization, classification and sparsification, namely, 
\begin{align*}
\operatorname{Loss}\left(x_{1}, x_{2}\right) = &\alpha \cdot \left(\operatorname{Loss}_{\text {rec }}\left(x_{1}\right) +  \operatorname{Loss}_{\text {rec }}\left(x_{2}\right)\right) +\\
&\beta \cdot \left(\operatorname{Loss}_{\text {var }}\left(x_{1}\right) +  \operatorname{Loss}_{\text {var }}\left(x_{2}\right)\right) +\\
&\gamma \cdot \left(\operatorname{Loss}_{\text {class }}\left(x_{1}\right) + \operatorname{Loss}_{\text {class }}\left(x_{2}\right)\right) +\\
&\, \theta \cdot \operatorname{Loss}_{\text {sparse }} \numberthis \label{eqn1}
\end{align*}
where
\begin{align*}
\operatorname{Loss}_{\text {rec }}(x) &=M S E(\mathbf{W} \odot x, \mathbf{W}^* \odot \tilde{x})\\
\operatorname{Loss}_{v a r}(x) &=D_{K L}\left(N\left(\mu, \sigma^2 I\right), N(0, I)\right)\\
\operatorname{Loss}_{\text {class }}(x) &=\log \operatorname{Loss}\left(p(x), \tilde{p}(x)\right)\\
\operatorname{Loss}_{\text {sparse }} &=\left\|\mathbf{W}\right\|_1 \numberthis \label{eqn2}
\end{align*}

The block-scheme of one training process is given by Algorithm \ref{alg:training}.
\begin{algorithm}[!htbp]
\caption{Training}
\label{alg:training}
  \Input{%
    \begin{tabular}[t]{ll}
      $\theta, \alpha, \beta, \gamma$ & weight of loss functions\\
      $argsOpt1$, $argsOpt2$ & parameters of Adam optimizers for VAE's \\
      $argsOptCls$ & parameters of Adam optimizers for Classifier \\
      $argsOptSL$ & parameters of Adam optimizers for SL\\
    \end{tabular}
    }
  \Output{%
    \begin{tabular}[t]{l}
      $\mathbf{W}$ - weights of SL\\
    \end{tabular}
    }
  \KwData{%
    \begin{tabular}[t]{ll}
      $[x_1,x_2]$ & list of pairs of equally seized batches sampled from both datasets \\ 
      & 20 000 elements, i.e epochs of training
    \end{tabular}
    }
  \Init{%
    \begin{tabular}[t]{l}
      SL, VAE1, VAE2, Classifier (initialize subnets) \\
      OptSL = AdamW($argsSL$) \\
      Opt1 = AdamW($argsOpt1$) \\
      Opt2 = AdamW($argsOpt2$) \\
      OptCls = AdamW($argsCls$)
    \end{tabular}
    }

  \BlankLine
    \For{$x_1,x_2 \in [x_1,x_2]$}
      {
      $sp(x_1), sp(x_2) = \mathbf{W} \odot x_1, \mathbf{W} \odot x_2$\\
      $\tilde{x}_1, \mu_1+\epsilon\sigma_1 = \text{VAE}1(sp(x_1))$\\
      $\tilde{x}_2, \mu_2+\epsilon\sigma_2 = \text{VAE}2(sp(x_2))$\\
      $\tilde{p}(x_1) = \text{Classifier}(\mu_1+\epsilon\sigma_1)$\\
      $\tilde{p}(x_2) = \text{Classifier}(\mu_2+\epsilon\sigma_2)$\\
      Calculate Loss as in \eqref{eqn1} \\
      Backpropagate Loss \\
      Step all optimizers \\
      }
\end{algorithm}
After the rough-tuning of the hyper-parameters from \cite{Uhler} along with the sparsity contribution and parameters for optimizers, we set the following hyper-parameters:
\begin{itemize}
    \item $\alpha, \, \beta, \, \gamma, \, \theta$ = 10, $10^{-4}$, 1, $10^{-4}$
    \item LR's for AdamW optimizer for VAE's, SL and Classifier = $10^{-4}$
    \item all other parameters for AdamW optimizer are set to Pytorch default 
\end{itemize}

\section{Data}
The data consists of mice bulk RNA sequences extracted from two different tissues:  spleen and liver, used as two different domains for the purposes of this paper. The mice that were exposed to Salmonella infection were monitored and categorized by health status 
as tolerant, resistant, susceptible, or delayed susceptible, the latter two being related to strains unifying the mice who died within 1 or 3 weeks respectively. 
In all our experiments we combine these latter two groups into one susceptible group. With this new labelling we have 31 and 9 tolerant samples, 27 and 7 resistant samples and 90 and 53 susceptible samples for spleen and liver domains respectively for all infected mice. Also, the data includes control samples representing the mice who had never been exposed to infection labeled as "never infected". This group accounts for 104 samples, with 93 samples from spleen and 11 samples from liver. Phenotypes of these samples are determined  based on their genetic strains.
Initial bulk RNA dataset was TMM-normalized, the outliers and duplicates have been detected and dropped out. Finally, the domain-specific data, i.e. combined RNA data for samples from spleen and combined RNA data for samples from liver, have been z-scored for each domain separately and filtered for common across tissues genes in all feature selection algorithms resulting in data samples consisting of 34,861 genes, i.e., the dimension
of the input space.

\section{Experiment} 
\begin{figure}[!htbp]
\centering
\subfigure[]{\includegraphics[width=.45\textwidth, height=0.35\textwidth]{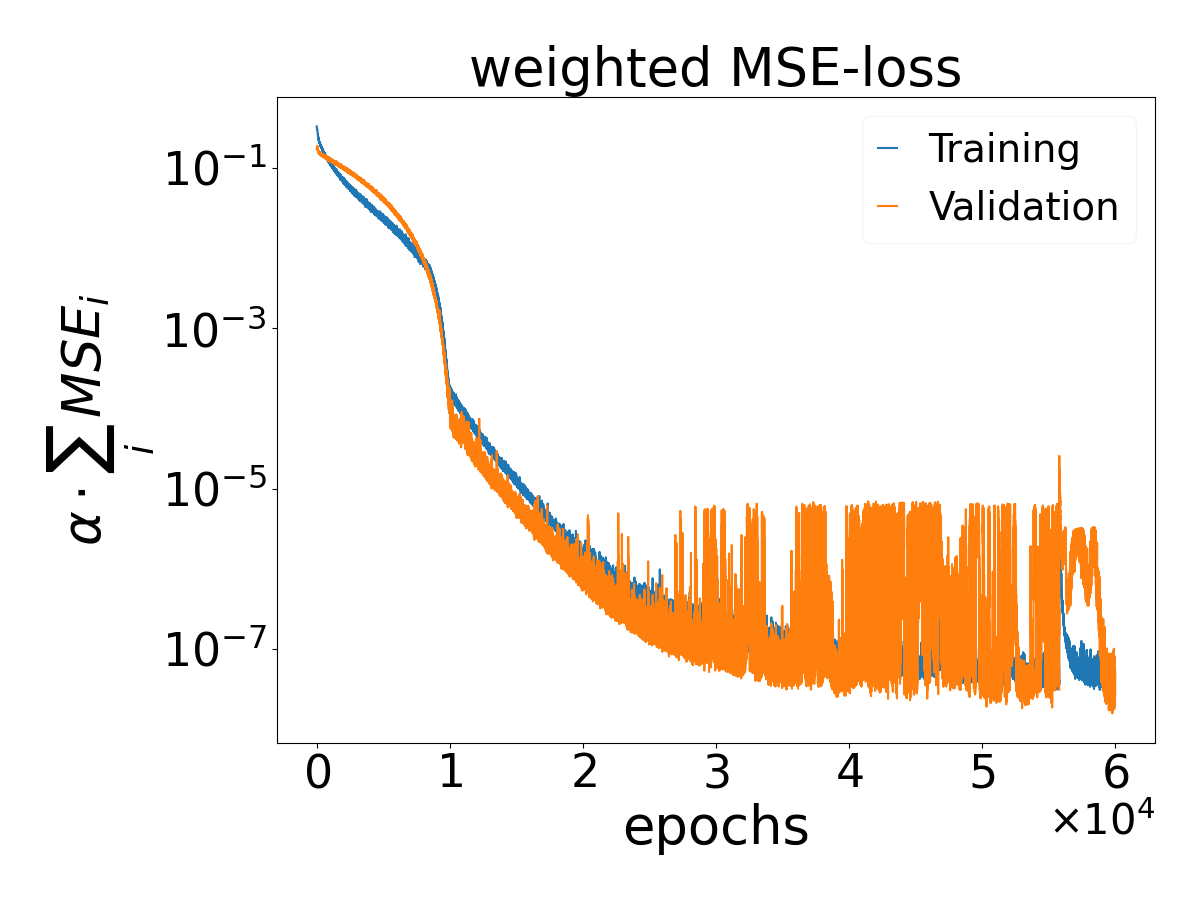}}
\subfigure[]{\includegraphics[width=.45\textwidth, height=0.35\textwidth]{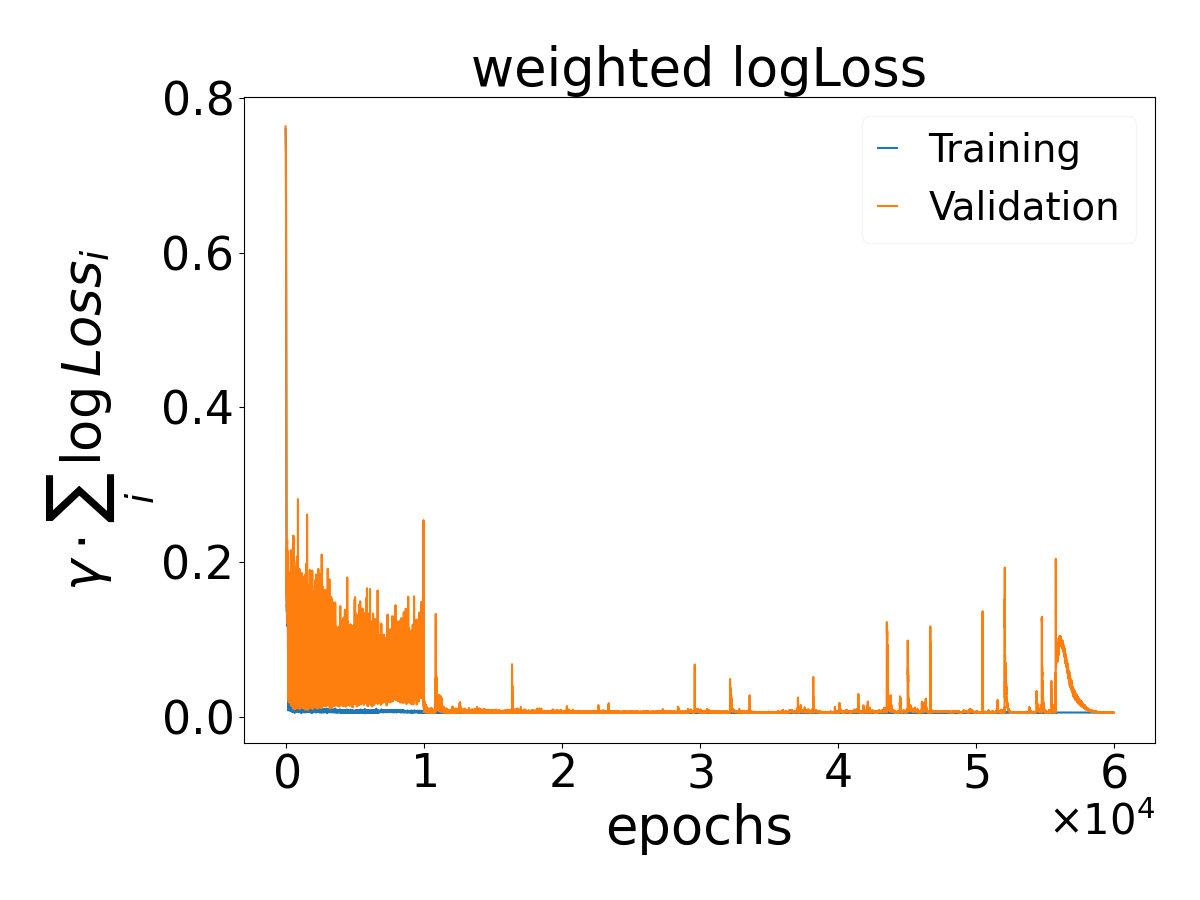}}
\subfigure[]{\includegraphics[width=.45\textwidth, height=0.35\textwidth]{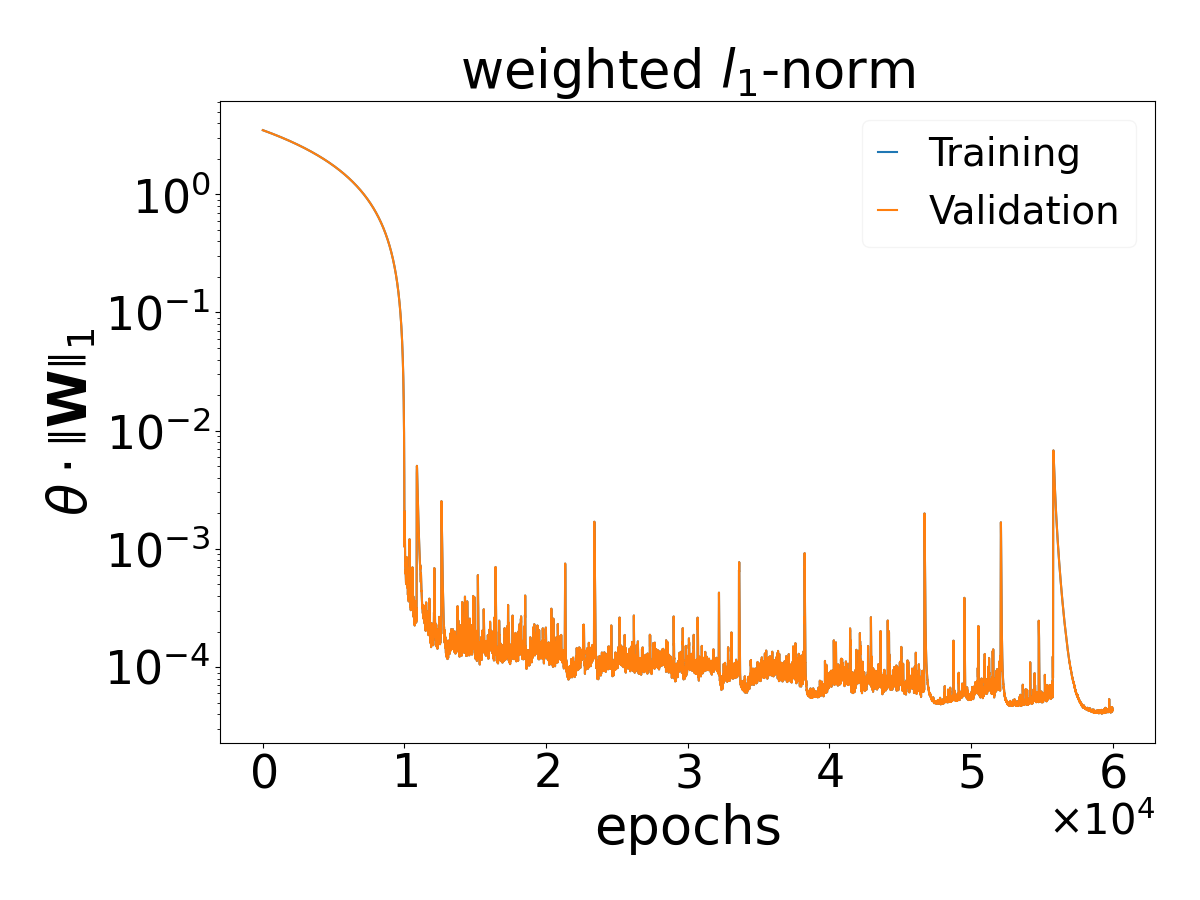}}
\subfigure[]{\includegraphics[width=.45\textwidth, height=0.35\textwidth]{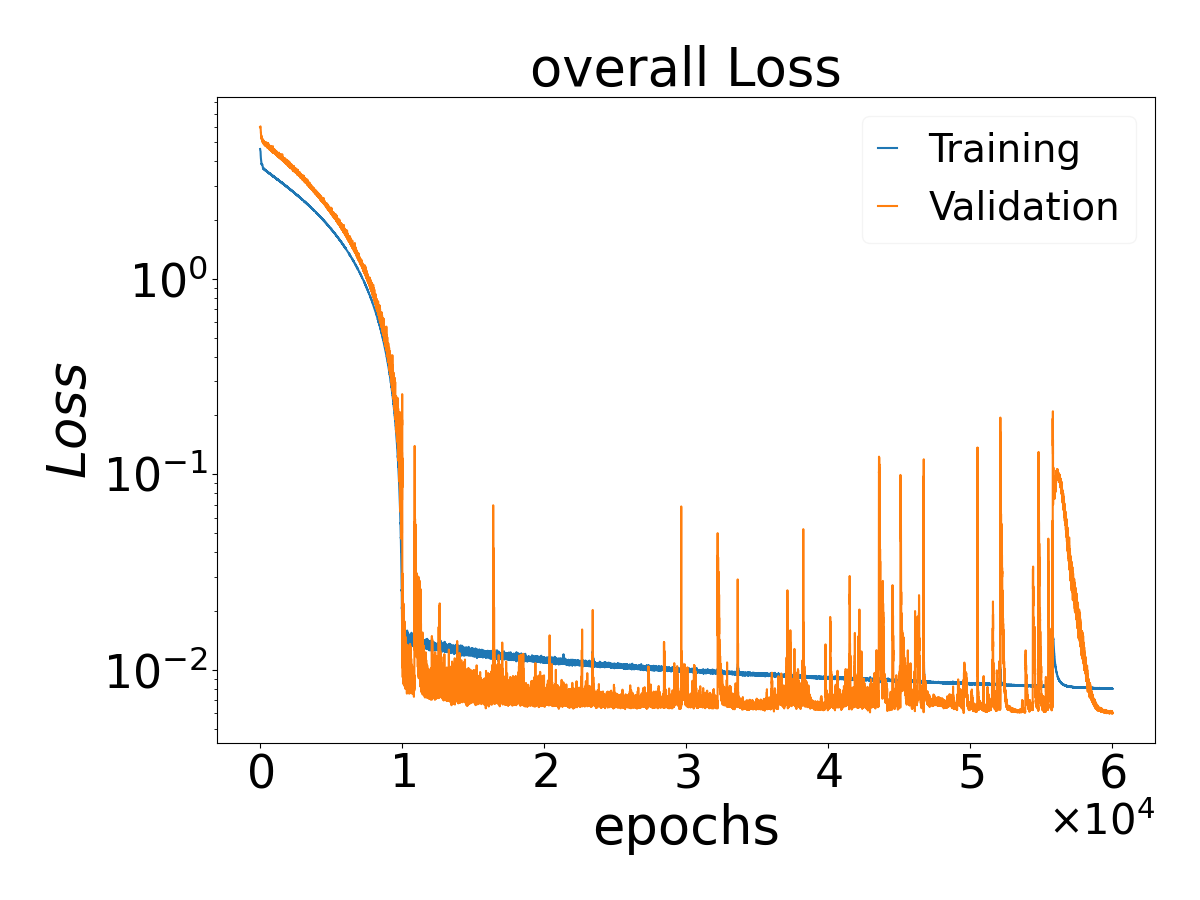}}
\caption{Weighted components of overall losses for phenotypes tolerant versus susceptible across domain experiment: (a) - reconstruction errors, (b) - classification errors, (c) - sparsity loss, (d) - overall loss.
}
\label{fig:errors1}
\end{figure}

\begin{figure}[!htbp]
\centering
\subfigure[]{\includegraphics[width=.45\textwidth, height=0.35\textwidth]{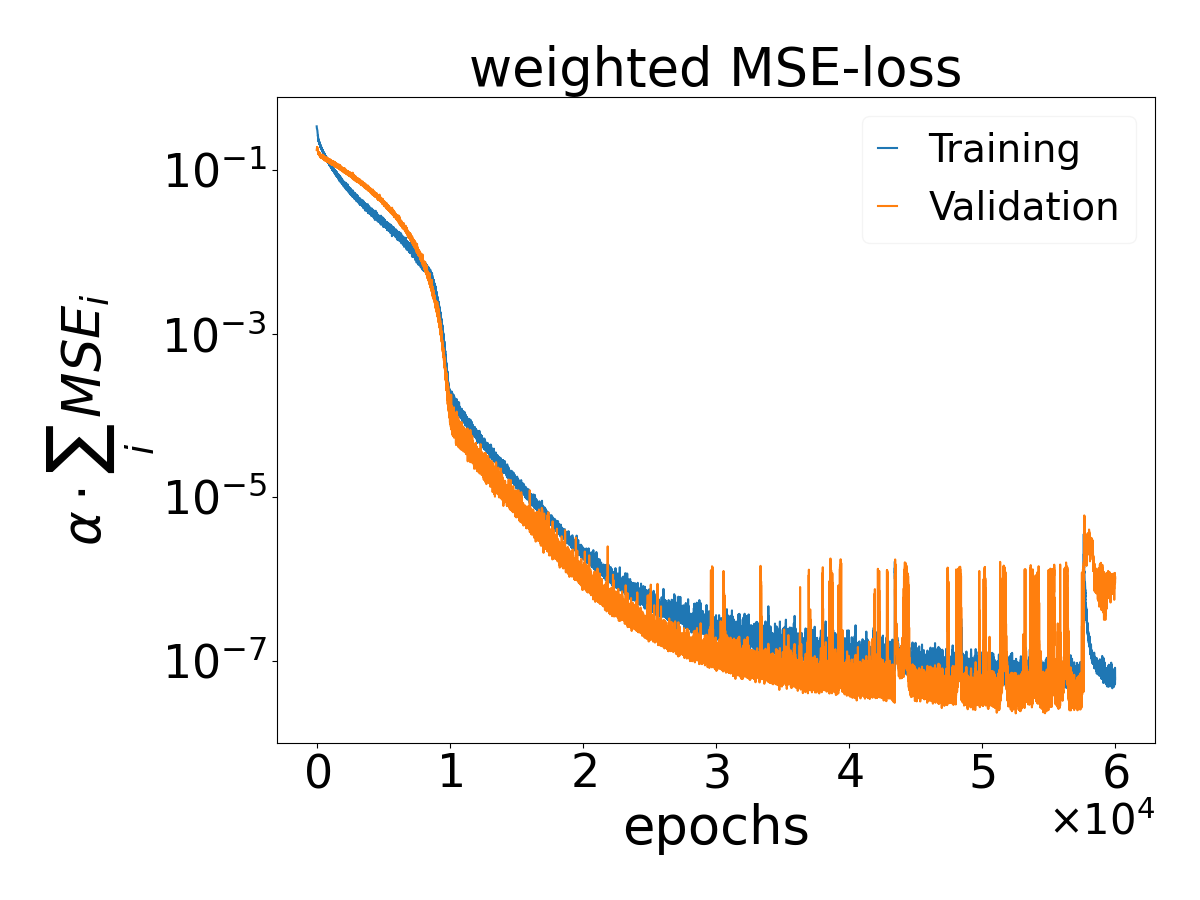}}
\subfigure[]{\includegraphics[width=.45\textwidth, height=0.35\textwidth]{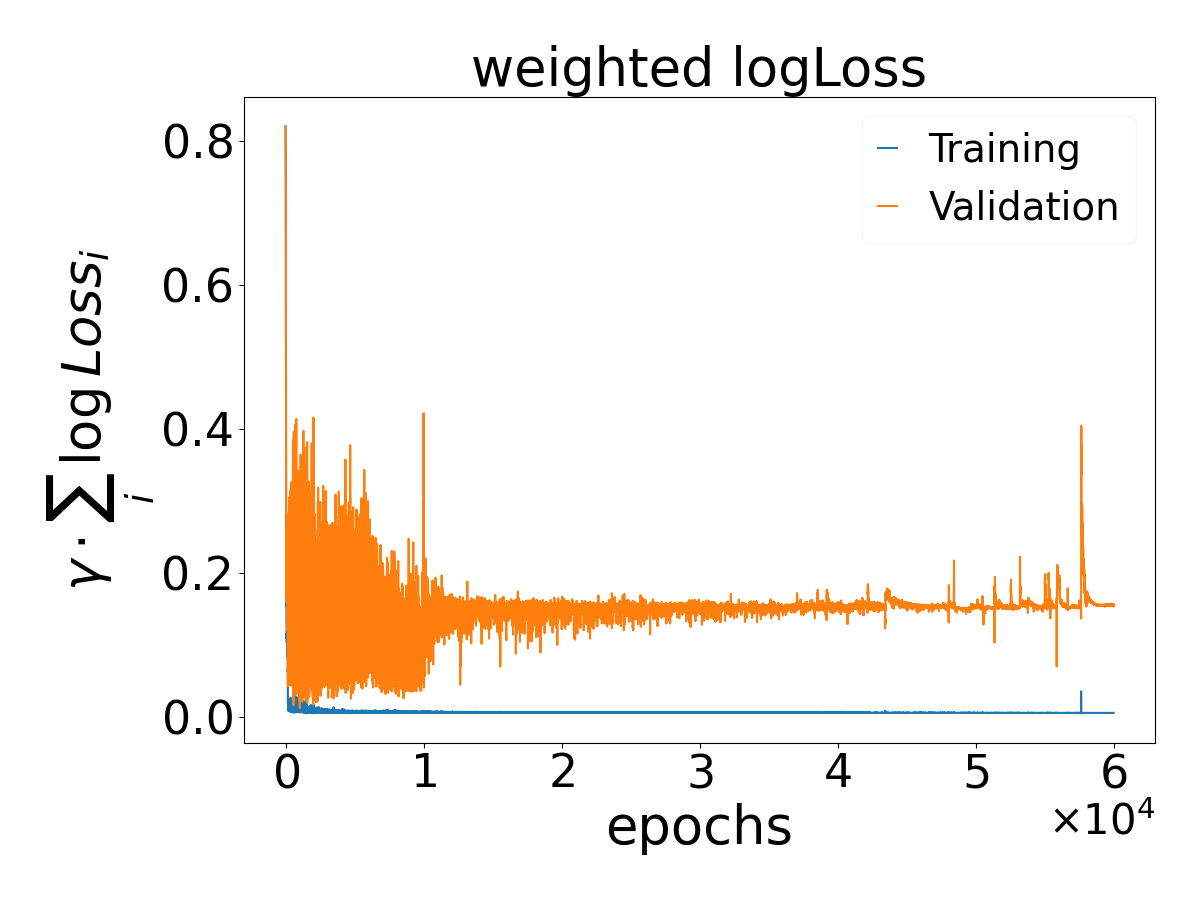}}
\subfigure[]{\includegraphics[width=.45\textwidth, height=0.35\textwidth]{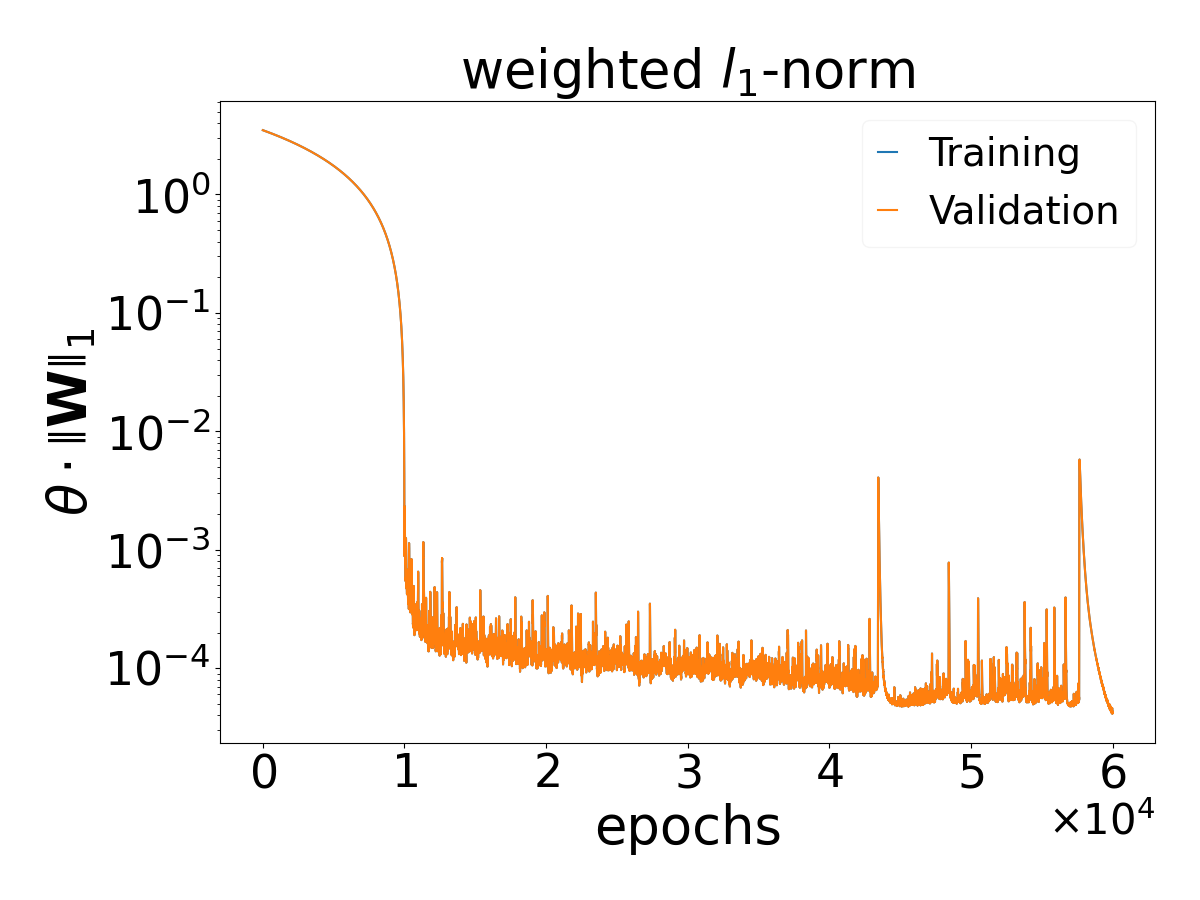}}
\subfigure[]{\includegraphics[width=.45\textwidth, height=0.35\textwidth]{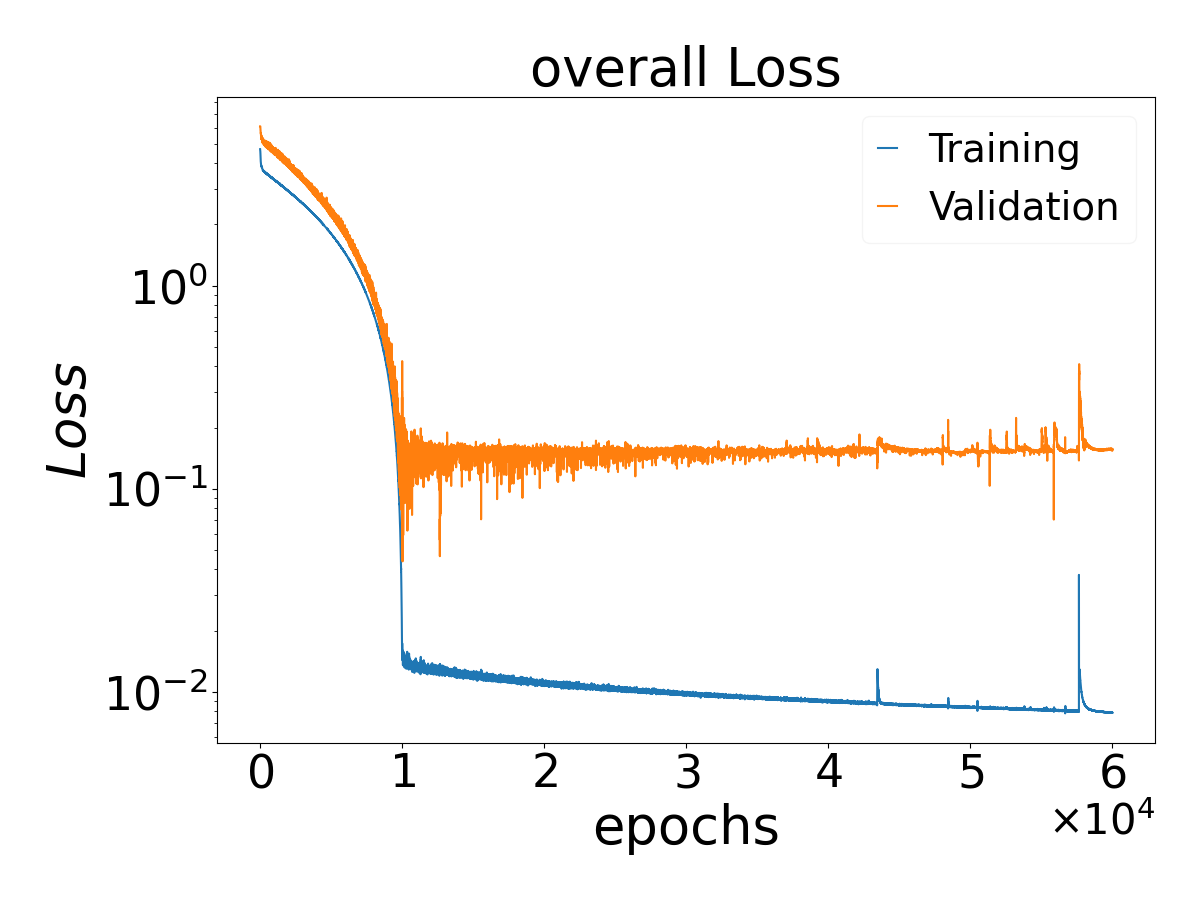}}
\caption{Weighted components of overall losses for phenotypes resistant versus susceptible across domain experiments: (a)  reconstruction error, (b)  classification error, (c)  sparsity loss, (d)  total loss.
}
\label{fig:errors2}
\end{figure}

\begin{figure}[!htbp]
\centering
\subfigure[]{\includegraphics[width=.45\textwidth, height=0.35\textwidth]{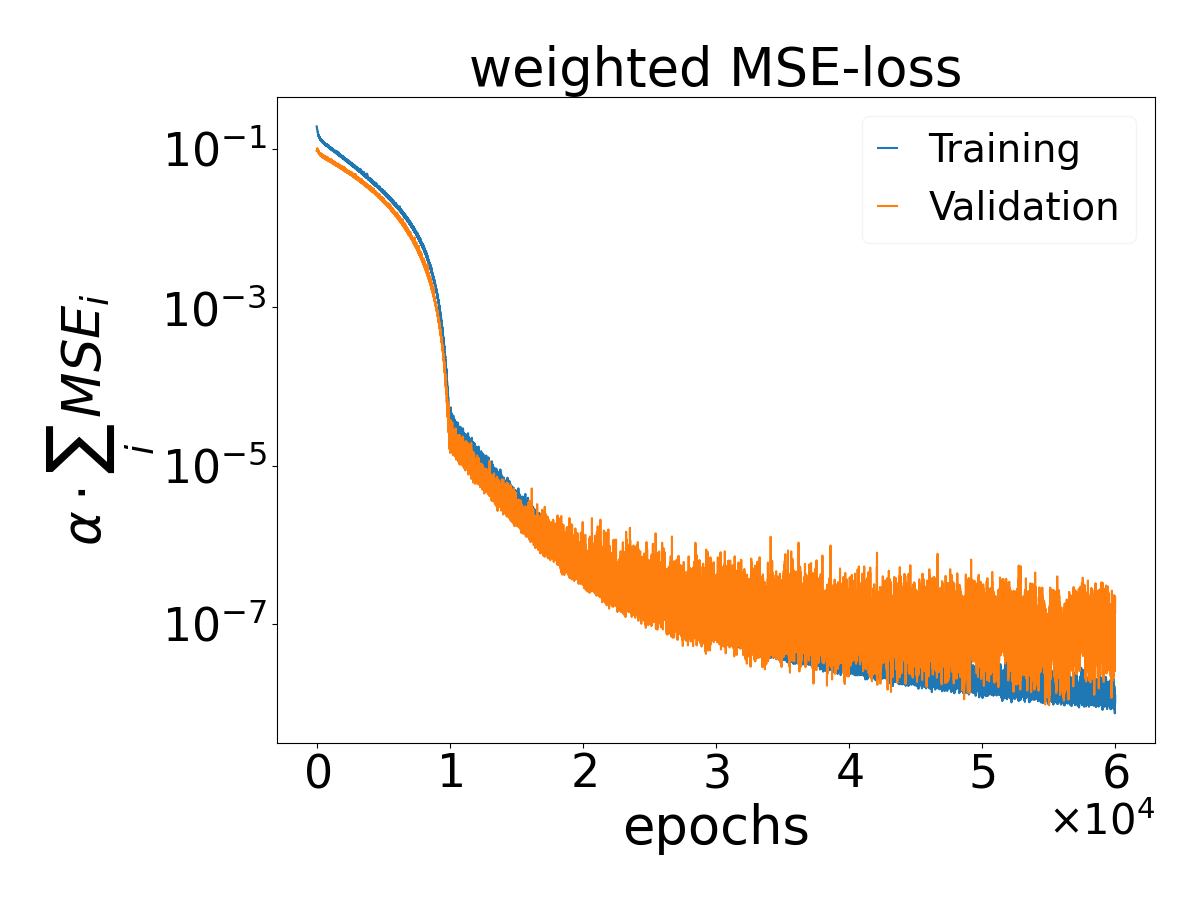}}
\subfigure[]{\includegraphics[width=.45\textwidth, height=0.35\textwidth]{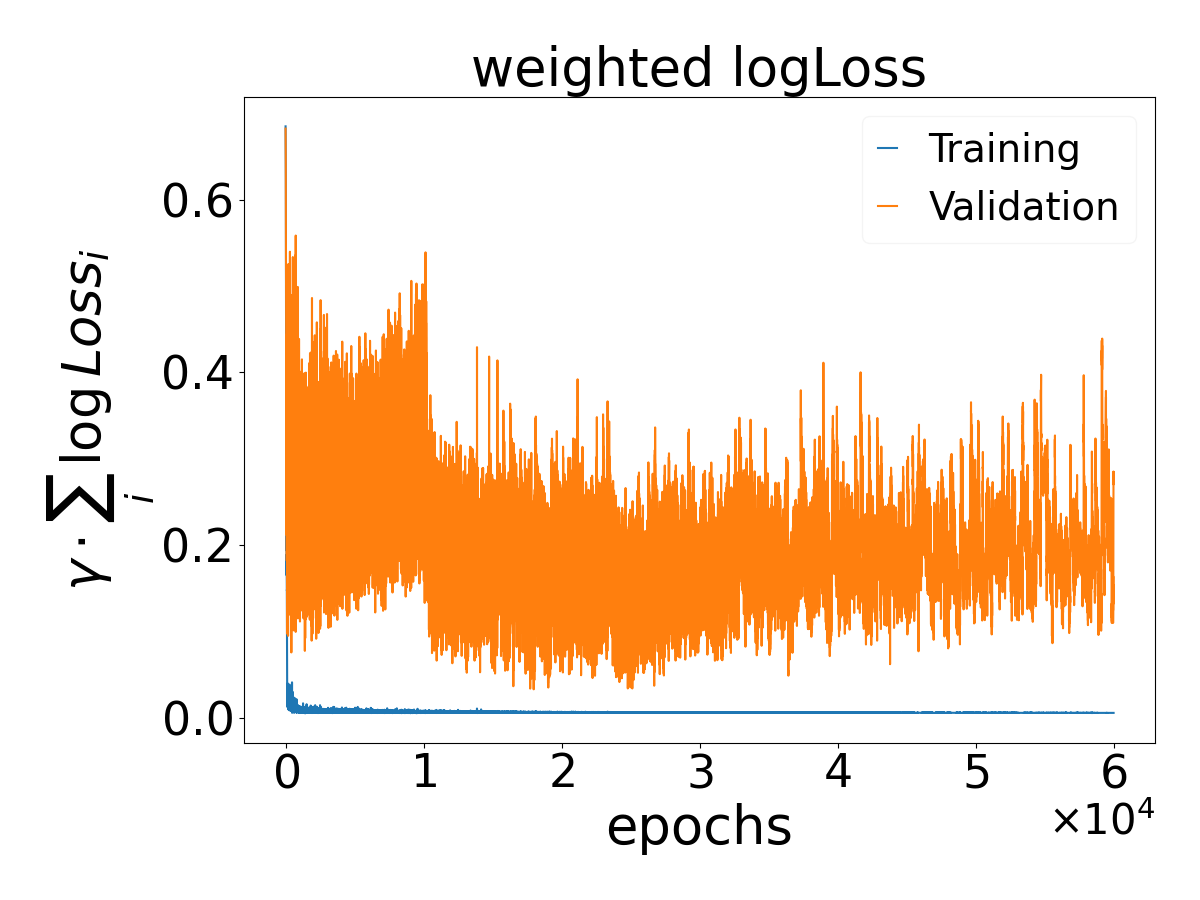}}
\subfigure[]{\includegraphics[width=.45\textwidth, height=0.35\textwidth]{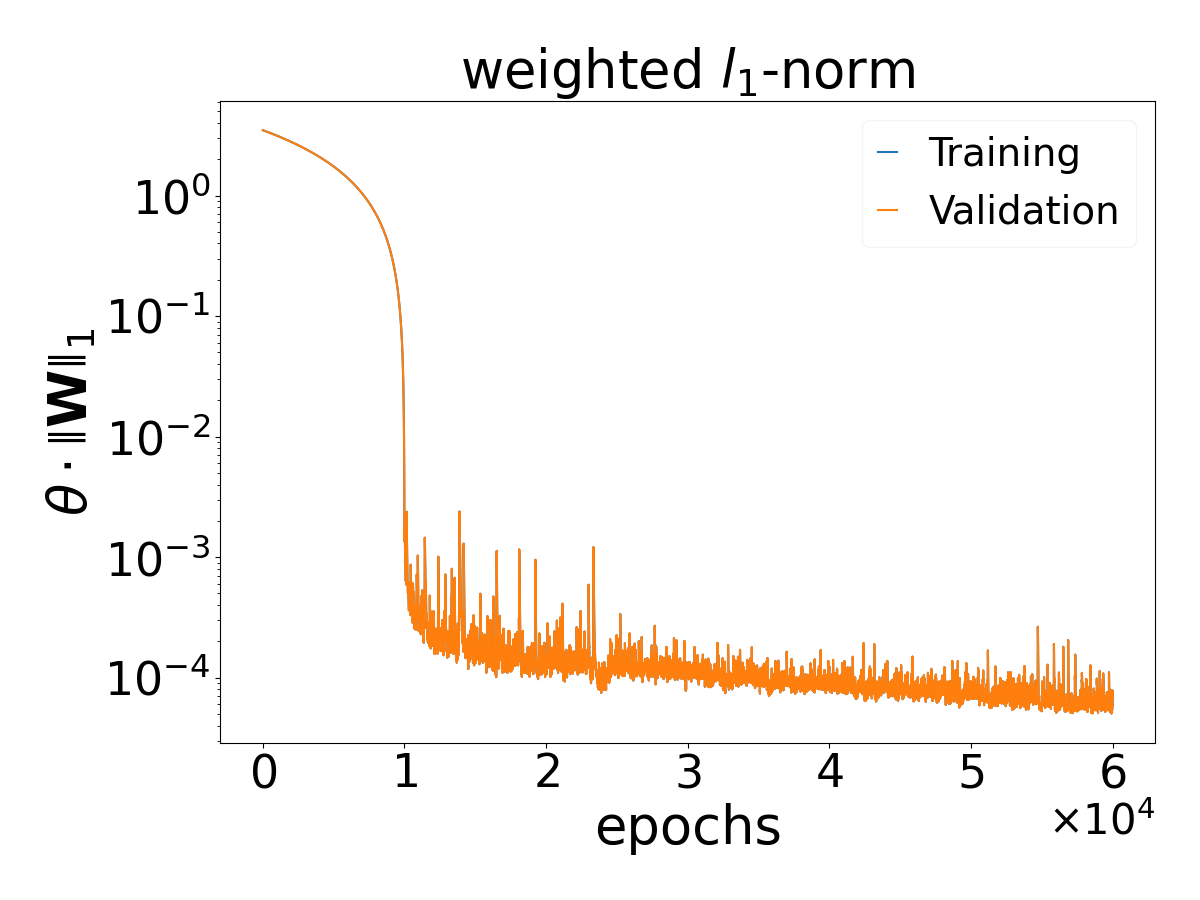}}
\subfigure[]{\includegraphics[width=.45\textwidth, height=0.35\textwidth]{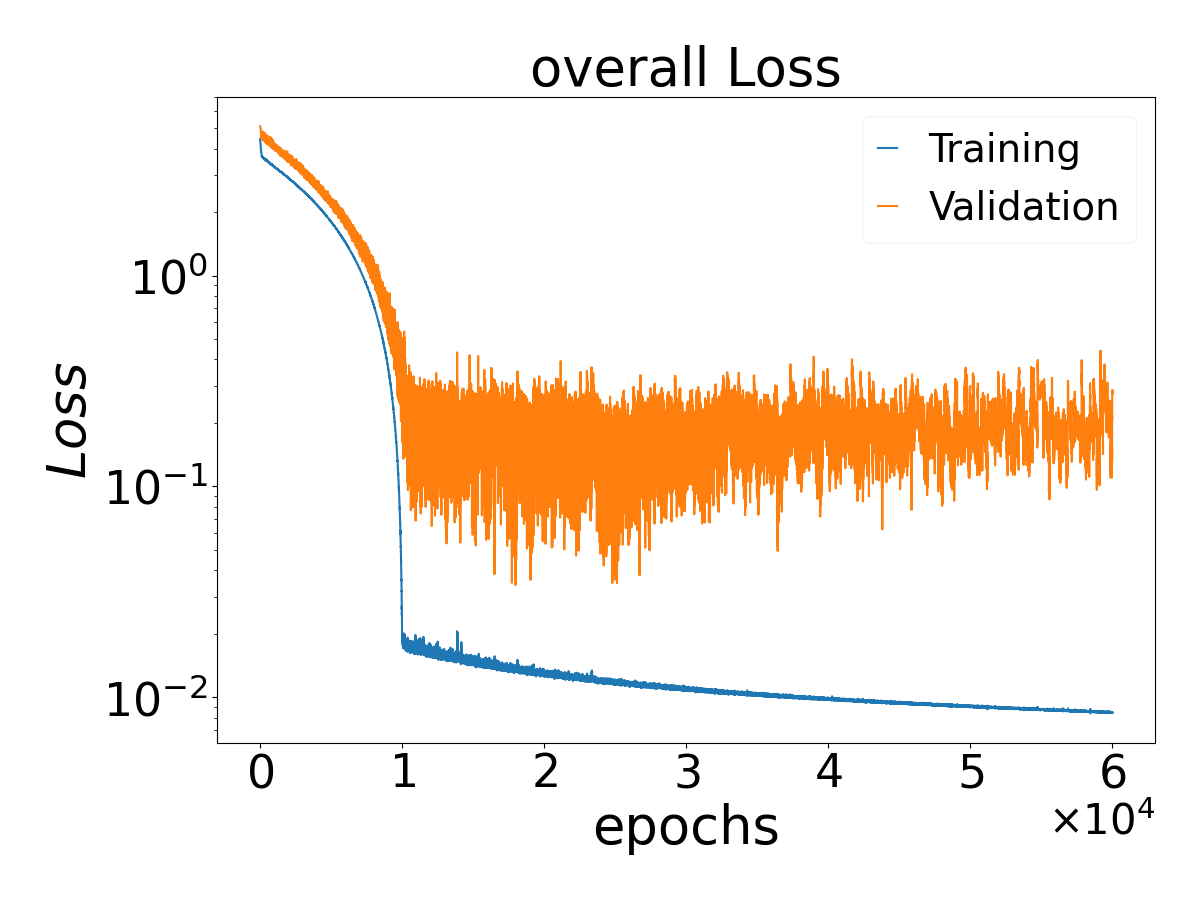}}
\caption{Weighted components of overall losses for infected mice versus never infected mice across domain experiment: (a) reconstruction error, (b) classification error, (c)  sparsity loss, (d) total loss.
}
\label{fig:errors3}
\end{figure}

We consider three distinct types of experiment.  In the first type the goal is to extract a small subset of features that discriminate among the phenotypes susceptible and tolerant. The second type extracts features discriminating among the phenotypes susceptible and resistant. The third type extracts the features discriminating between infected and never infected mice. The exact Python code with the training models and post-processing utilities is available at \cite{Code}.
With a use of Ray package \cite{Ray} the experiment was run on 16 V100 GPU's in a multiprocessing mode for 10 different random samplings of 85$\%$ of data, with 90 different weights initializations for each, summing up to 900 runs. You can see the typical evolution of training process including 60,000 epochs for all three experiments in Figures \ref{fig:errors1}, \ref{fig:errors2}, \ref{fig:errors3}. 

\begin{figure}[!htbp]
\centering
\subfigure[]{\includegraphics[width=.32\textwidth, height=0.33\textwidth]{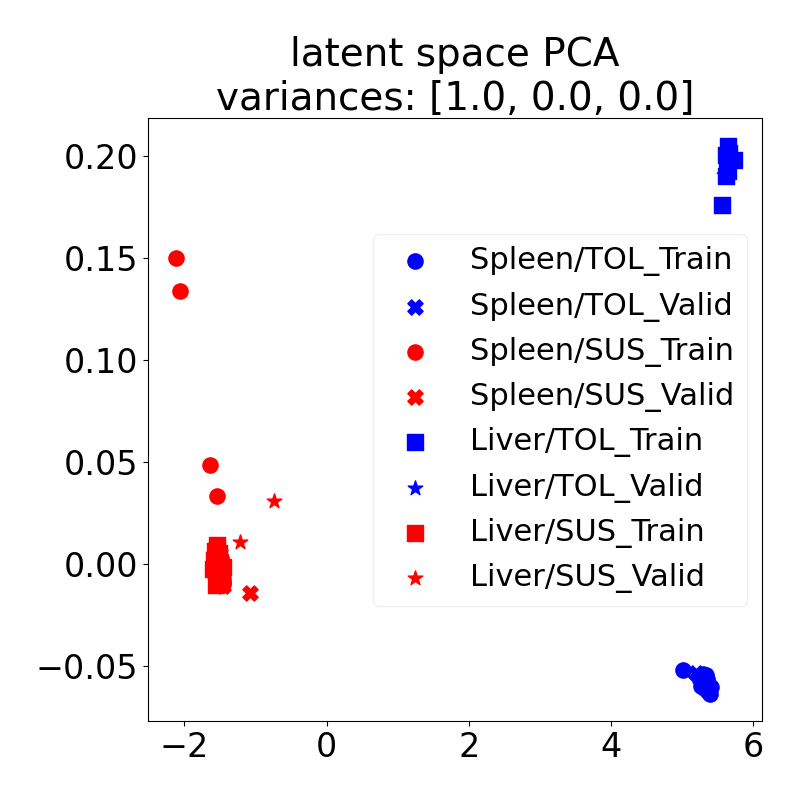}}
\subfigure[]{\includegraphics[width=.32\textwidth, height=0.33\textwidth]{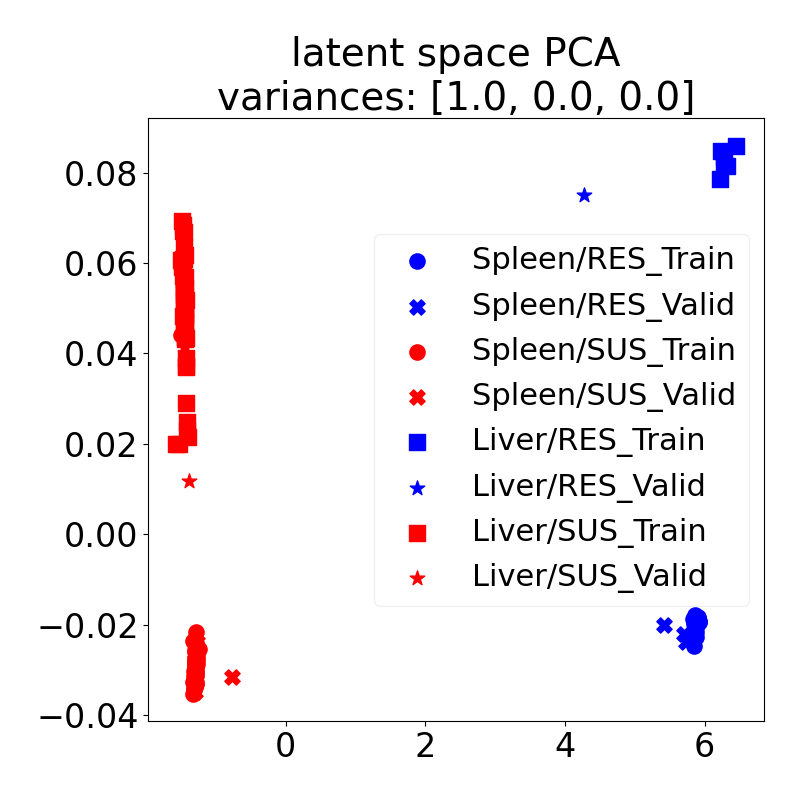}}
\subfigure[]{\includegraphics[width=.33\textwidth, height=0.33\textwidth]{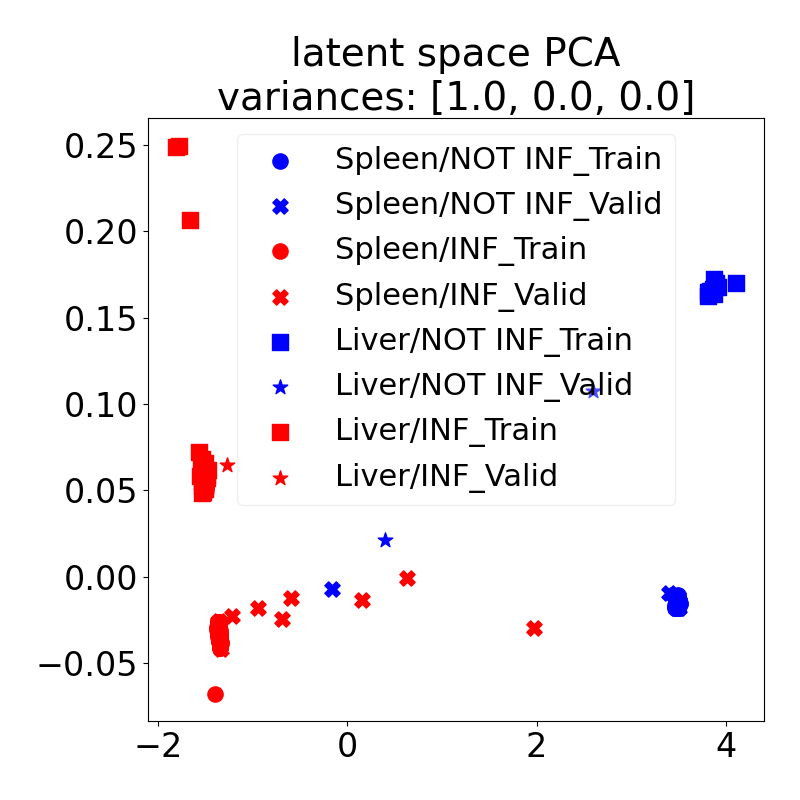}}
\caption{Latent space PCA's for all three across domain experiments: (a) phenotypes tolerant (TOL) versus susceptible (SUS), (b)  phenotypes resistant (RES) versus susceptible (SUS), (c)  infected (INF) mice versus never infected mice (NOT INF).
}
\label{fig:embedding}
\end{figure}

The number of epochs was fixed at 20,000 based on the indication of the flattening of the sparsity curves as long as it doesn't effect the accuracy of classification and the reconstruction loss is relatively small. Again, the reconstruction is not in the primary focus of this method, i.e, it wasn't the major task to consider when deciding on the number of epochs, especially, since the required for the across-domain classification alignment in the latent space was typically achieved even after 20,000 epochs, see the Figure \ref{fig:embedding}. The visualization of loss indicates that further sparsification slightly decreases the classification accuracy for two out of three experiments, by limiting the number of epochs we also prevent this long-run negative effect. The PCA images for all figures represent the PCA of combined representations of both domains in their bottleneck layers, and indicate a proper clustering and separability across domains, i.e., good alignment in latent space.

\section{Results}
\begin{figure}[!htbp]
\centering
\subfigure[]{\includegraphics[width=.32\textwidth, height=0.22\textwidth]{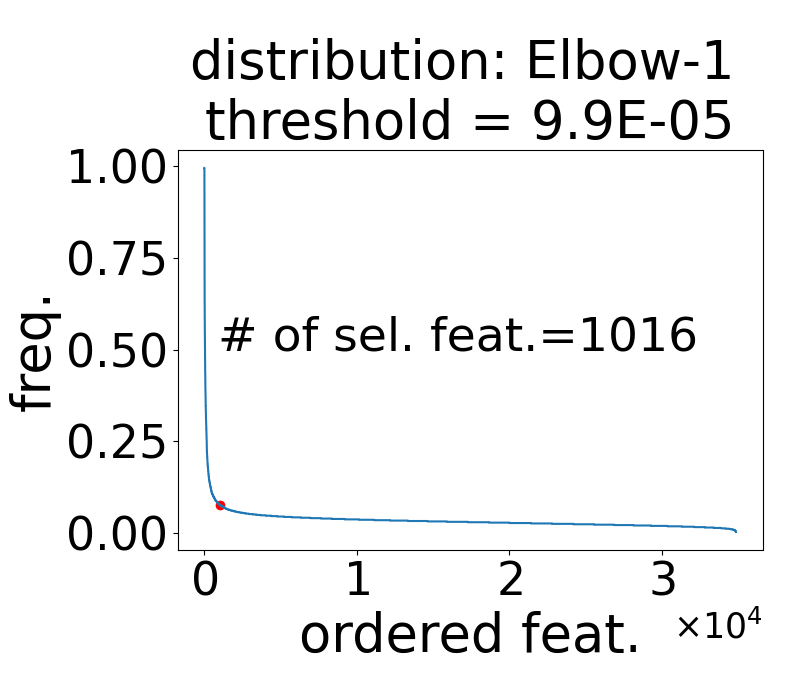}}
\subfigure[]{\includegraphics[width=.32\textwidth, height=0.22\textwidth]{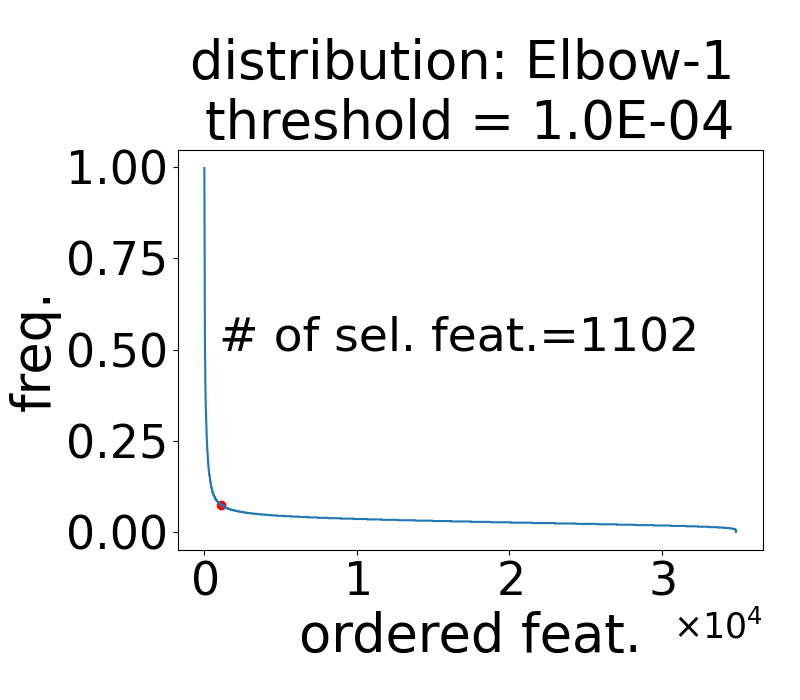}}
\subfigure[]{\includegraphics[width=.32\textwidth, height=0.22\textwidth]{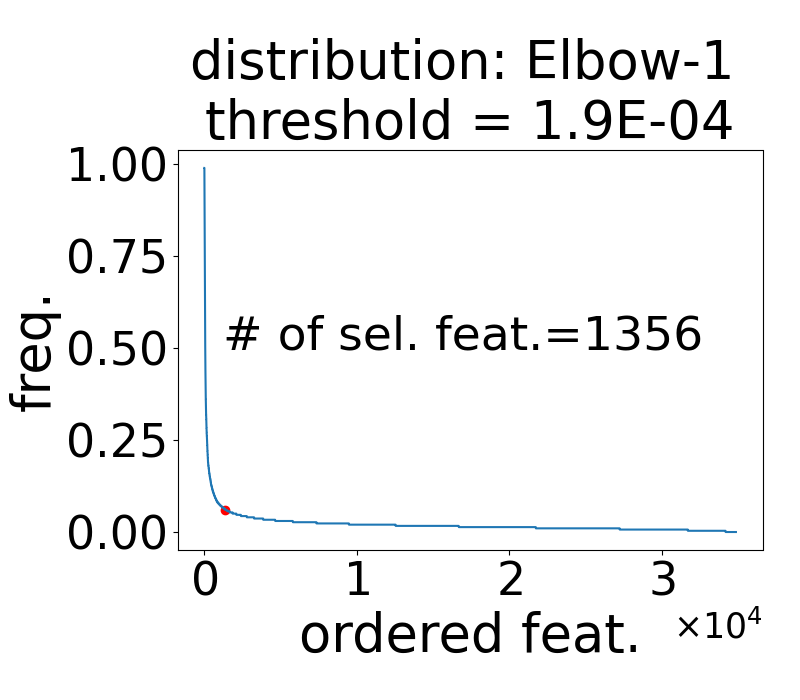}}
\subfigure[]{\includegraphics[width=.32\textwidth, height=0.22\textwidth]{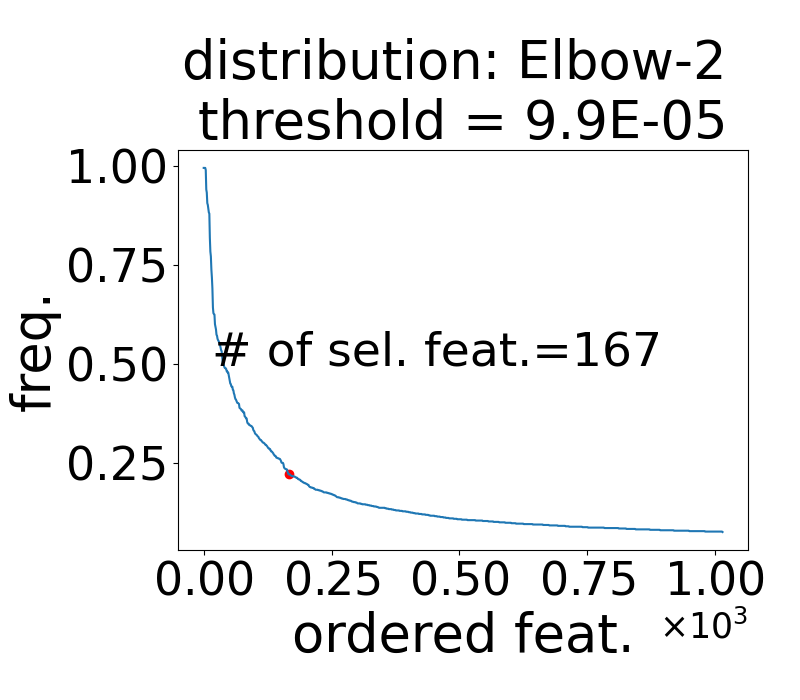}}
\subfigure[]{\includegraphics[width=.32\textwidth, height=0.22\textwidth]{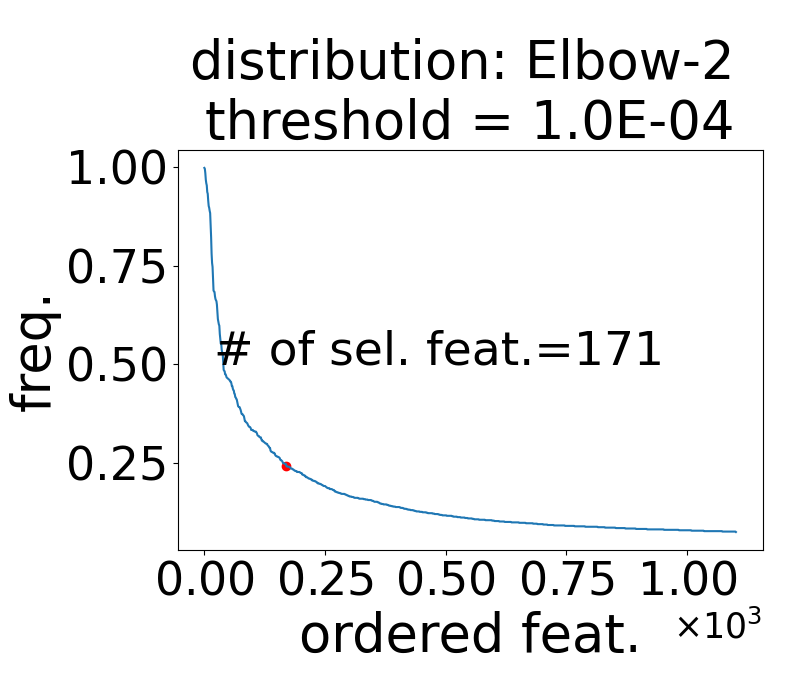}}
\subfigure[]{\includegraphics[width=.32\textwidth, height=0.22\textwidth]{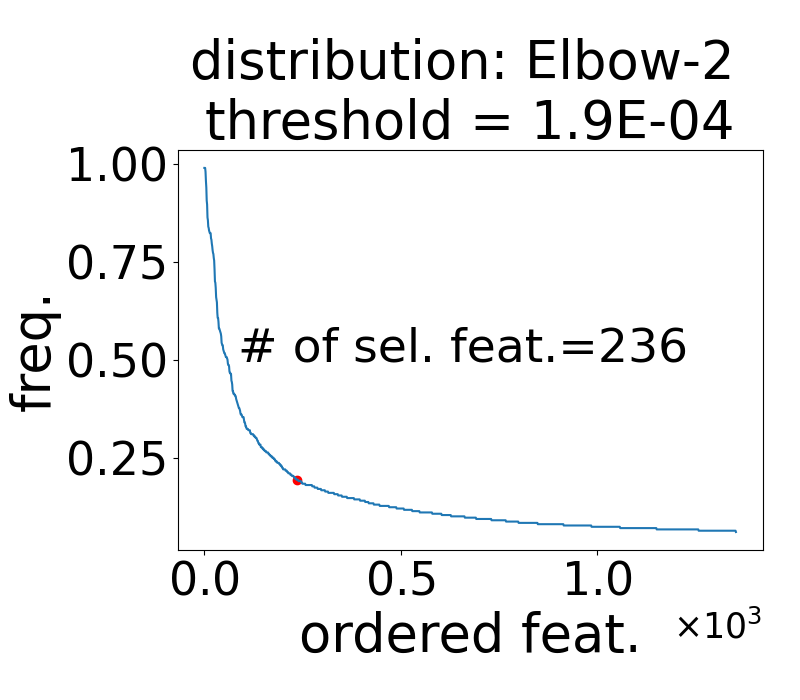}}
\caption{Results from the across domain (MDMT) experiments. The features selected are ordered by frequency for 3 experiments: panels (a) and (d) correspond to the phenotypes tolerant versus susceptible; panels (b) and (e) correspond to resistant versus susceptible; panels (c) and (f) correspond to infected versus never infected.}
\label{fig:freqs}
\end{figure}

The post-processing of the sparsity layer weights was conducted in a same way for all 3 experiments. Firstly, all the weights across 900 runs have been aggregated, normalized and the Elbow Method was applied to find a threshold, and later for each run the weights below the threshold were set to zero. At the next step we calculated the frequencies of features appearing in subsets of features with non-zero weights across all runs. The resulting distributions of frequencies are shown in Figure \ref{fig:freqs}, along with resulting number of features selected in two consecutive steps by Elbow Method.

\begin{figure}[!htbp]
\centering
\subfigure[]{\includegraphics[width=.32\textwidth, height=0.2\textwidth]{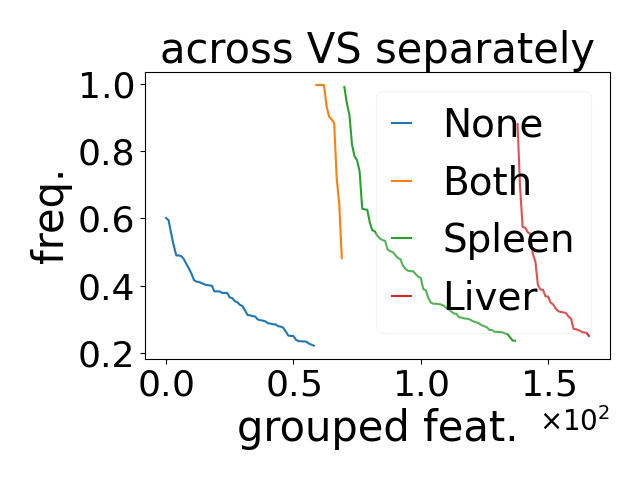}}
\subfigure[]{\includegraphics[width=.32\textwidth, height=0.2\textwidth]{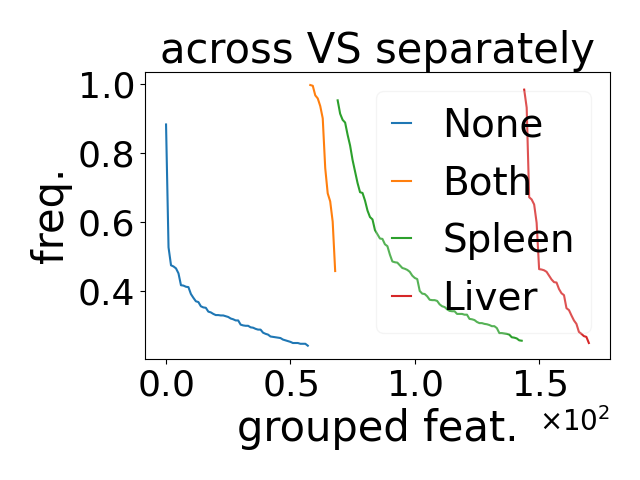}}
\subfigure[]{\includegraphics[width=.32\textwidth, height=0.2\textwidth]{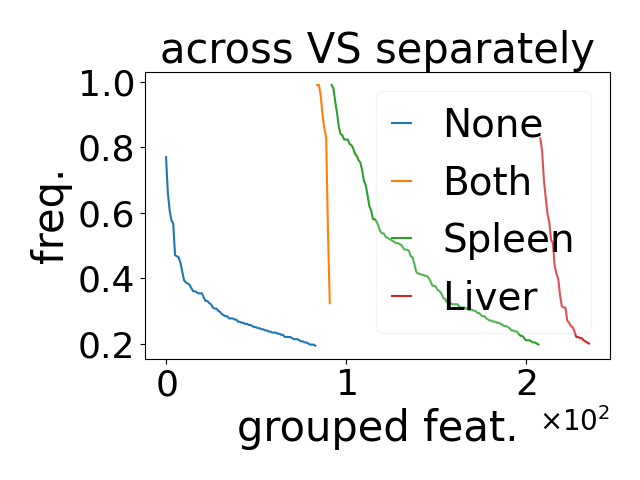}}
\subfigure[]{\includegraphics[width=.32\textwidth, height=0.2\textwidth]{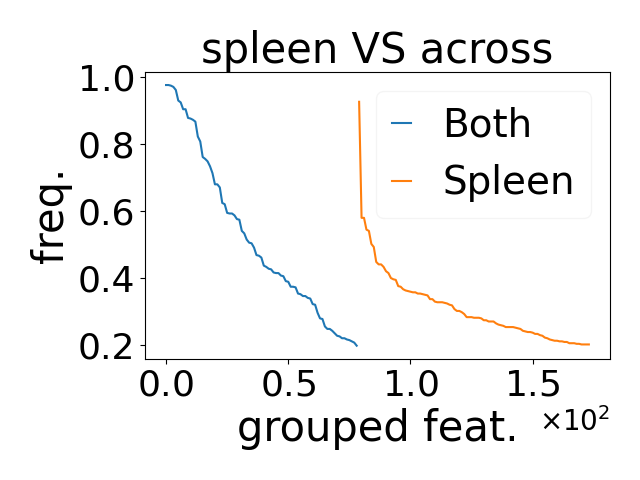}}
\subfigure[]{\includegraphics[width=.32\textwidth, height=0.2\textwidth]{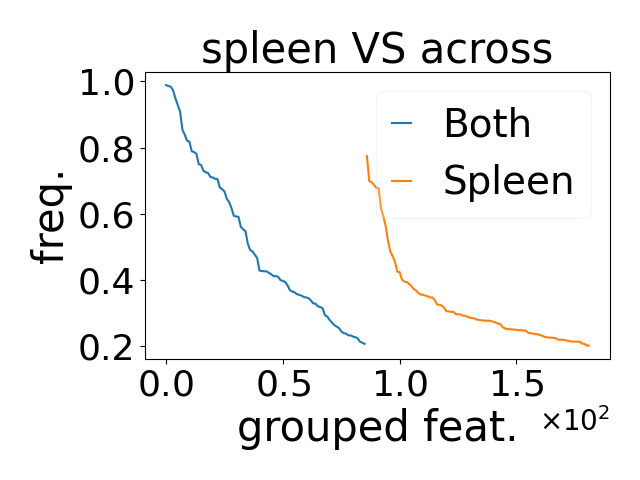}}
\subfigure[]{\includegraphics[width=.32\textwidth, height=0.2\textwidth]{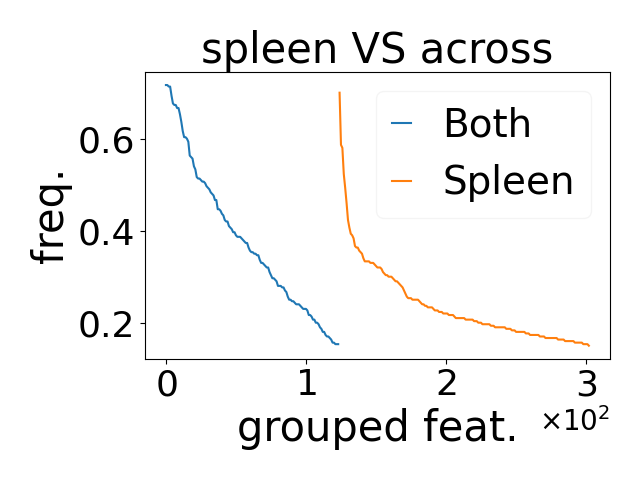}}
\subfigure[]{\includegraphics[width=.32\textwidth, height=0.2\textwidth]{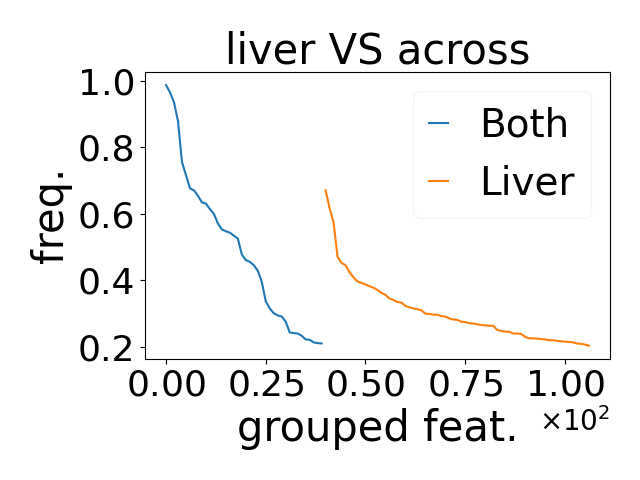}}
\subfigure[]{\includegraphics[width=.32\textwidth, height=0.2\textwidth]{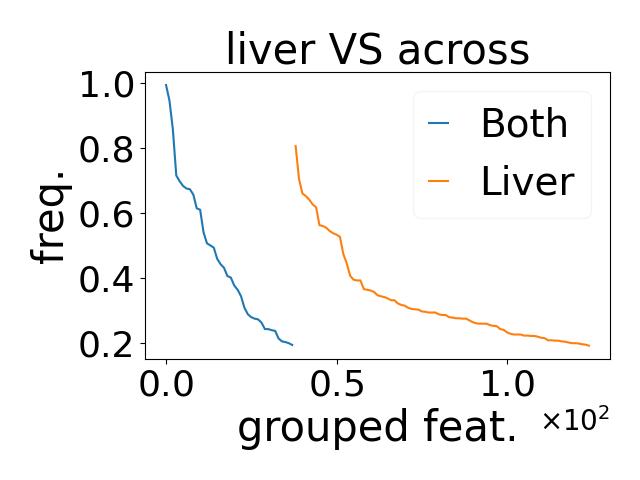}}
\subfigure[]{\includegraphics[width=.32\textwidth, height=0.2\textwidth]{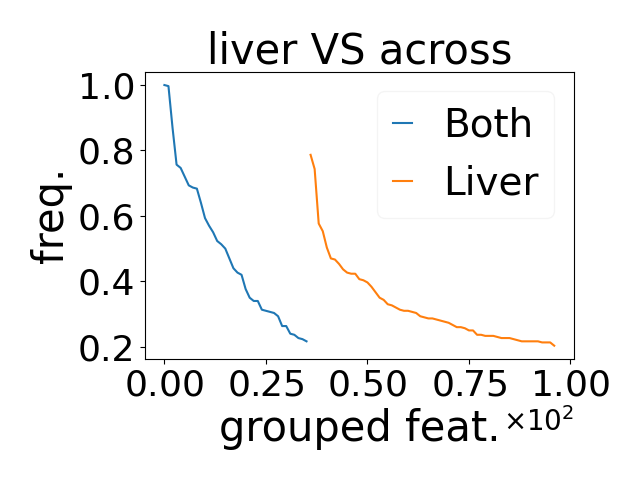}}
\caption{Grouped by overlapping distribution of features for all experiments. Columns are associated with experiments, i.e. 1st - tolerant versus susceptible, 2nd - resistant versus susceptible, 3rd - infected versus never infected. (a),(b),(c)  distributions of features extracted in across domains experiments; (d),(e),(f)  features extracted in separate spleen experiment, features are grouped by overlapping with features extracted in respective across domains experiment: "Both" - overlapping, "Spleen" - not overlapping; (g),(h),(i)  features extracted in separate liver experiment, features are  grouped by overlapping with features extracted in respective across domains experiment: "Both" - overlapping, "Liver" - not overlapping.}
\label{fig:overlapSmall}
\end{figure}

Importantly, 
in addition to the cross domain learning, we also 
selected features for each domain separately for all 
three experiments.
This allows us to evaluate the distinct characteristics
of single domain and multi-domain alignment for feature extraction. 
In Figure
\ref{fig:overlapSmall} panels (a), (b), and (c), we can see the distribution of features for all three across domain experiments grouped by overlapping with features selected in one-domain experiments.
"Both" features is a subset of these features that also appear in both the spleen  and liver domains. "None" refers to the features that are the features that only appear in across domain results. "Spleen" are the features appearing in both across domains and spleen domain results but not in the liver domain results. Finally, "Liver" features are the features appearing in both across domains and liver domain results but not in the spleen domain results. We can see that for all the experiments apart from the features captured from domain-specific experiments some new highly-weighted features were captured in across domain experiments (the blue line denoting "None"). These features that are only present in the across domain experiment
reflect the new information being captured by the proposed method.  These correspond to biomarkers that we suspect
have a potentially unique role in the host response to infection.

Figures \ref{fig:overlapSmall}, panels (d),(g),(e),(h),(f), and (i) show the distributions of features extracted in one domain experiments grouped by overlapping with respective across domains results. Even though the majority of the most discriminative features extracted in separate experiments are also extracted in across domain experiments, some highly-weighted features extracted from one domain experiments are apparently not captured in respective across domains experiments.
These results might indicate that the improvement in robustness is needed, but at the same time they may contain biological insights about the across domain importance of some features that couldn't be obtained from studying only one domain for Salmonella infection or about the difference in manifestations of infection in different tissues. Based on the results from additional experiments with other
datasets
most likely the latter is true.


Returning to Figure \ref{fig:embedding},  we see  the
results of the embedding of the sparsified input in a two-dimensional latent space using the features found in the across domain 
architecture for all three experiments.  
We observe that the tolerant and resistant versus
susceptible experiments give excellent classification
on test data in the latent space.  In contrast, infected versus control mice are not as easily discriminated.

\section{Conclusions}
This paper proposes a novel architecture
for multi-domain, multi-task
feature selection. 
This area of research has a relatively small
literature, possibly 
because of the complexity associated
with simultaneously
exploring multiple incommensurate
measurement domains.
The proposed approach leverages prior art
in multi-domain learning 
while adding a masked feature selection
approach that serves to identify biologically
relevant aspects of the host immune
response to infection.  We demonstrate that
the demands of the  MDMT problem
formulation can be successfully addressed with
the proposed architecture. Further, 
the application of the approach
leads to the discovery
of novel biomarkers whose signals appear to be amplified by the multi-domain approach; indeed, a fraction of these biomarkers do not appear in either  single experiment.  
Hence, this approach holds the promise of
generating new biological insights that might
go undetected using single domain methodologies.
Additionally, we observed the the MDMT features
provide excellent classification results between susceptible and tolerant or resistant phenotypes.

There are several possible modifications of our method. In this paper the optimization problem descends in the direction of the gradient 
of weighted
objectives.  There is growing evidence that
isolating descent directions to improve
individual cost functions may lead to improved solutions.  Additionally, alternative data reduction
and reconstruction mappings could be
explored reflecting recent developments
in deep neural networks including graph
convolutional neural networks or
transformers.   This preliminary work is
focused on algorithm development; we propose 
to explore the biological ramifications of the
biomarkers in future work.

We believe  further development in these directions could lead to even more efficient methods and provide biologists with a new perspective on understanding the evolution of infections in tissues.

\medskip
{\bf Acknowledgements:}
    We would like to thank Helene Andrews-Polymenis and David Threadgill for providing the data for this investigation.


\medskip



\begin{thebibliography}{8}

\bibitem{filters}
Blum, A. L., Langley, P.: Selection of relevant features and examples in machine learning, Artif. Intell., \textbf{97}, 1, pp. 245--271 (1997)

\bibitem{wrappers}
John, G.H., Kohavi, R., Pfleger, K.: Irrelevant features and the subset selection problem, Proc. 11th Int. Conf. Mach. Learn., pp. 121--129 (1994)

\bibitem{AEUFS}
Han, K., Wang, Y., Zhang, C., Li, Ch., Xu, C.: AutoEncoder Inspired Unsupervised Feature Selection, ArXiv, \textbf{1710.08310}, \url{https://arxiv.org/abs/1710.08310} (2017)

\bibitem{CAE}
Fatih Balin, M., Abid, A., Zou, J.Y.: Concrete autoencoders: Differentiable feature selection and reconstruction, ICML (2019)

\bibitem{CAEbase}
Maddison, C. J., Mnih, A., Teh, Y. W.: The Concrete Distribution: A Continuous Relaxation of Discrete Random Variables, ArXiv, \textbf{1611.00712}, \url{https://arxiv.org/abs/1611.00712} (2016)

\bibitem{FsNet}
Singh, D., E., Yamada, M.: Fsnet: Feature selection network on high-dimensional biological data. ArXiv, \textbf{2001.08322}, \url{https://arxiv.org/abs/2001.08322}, (2020)

\bibitem{DFS}
 Li, Y., Chen, C., Wasserman, W.: Deep feature selection: theory and application to identify enhancers and promoters. Journal of Computational Biology,  \textbf{23}(5), pp. 322–-336 (2016)

\bibitem{SINNHNRC}
Feng J., Simon, N.: Sparse-Input Neural Networks for High-dimensional Nonparametric
Regression and Classification, ArXiv, \textbf{1711.07592}, \url{https://arxiv.org/abs/1711.07592} (2017)

\bibitem{StackedAE}
Hinton, G., Salakhutdinov, R.: Reducing the dimensionality of data with neural
networks. Science, \textbf{313}, pp. 504--507 (2006)

\bibitem{ContractiveAE}
Salah, R., Pascal, V., Xavier, M., Xavier, G, Yoshua, B.: Contractive Auto-Encoders: Explicit Invariance During Feature Extraction, Proceedings of the 28th International Conference on Machine Learning, ICML '11, pp. 833--840, New York (2011)

\bibitem{DBN}
Hinton, G., Osindero, S., Teh, Y.: A fast learning algorithm for deep belief nets.
Neural Computation, \textbf{18}, pp. 1527--1554 (2006)

\bibitem{STG}
Yamada, Y., Lindenbaum, O., Negahban, S., Kluger., Y.: Feature selection using stochastic gates. International Conference on Machine Learning, PMLR, \textbf{119}, pp. 10648--10659, (2020)

\bibitem{LassoNet}
Lemhadri, I., Ruan, F., Abraham, L., Tibshirani, R.: Lassonet: A neural network with feature
sparsity. Journal of Machine Learning Research, \textbf{22}(127), pp. 1--29, (2021)

\bibitem{Lasso}
Tibshirani, R.: Regression shrinkage and selection via the lasso. Journal of the Royal Statistical Society, Series B, \textbf{58}, pp. 267--288, (1996)
\bibitem{ElasticNet}
Zou, H., Hastie, T.: Regularization and variable selection via the elastic net., J. R.
Stat. Soc. Series B Stat. Methodol. \textbf{67}(2), pp. 301--320 (2005)
\bibitem{FusedLasso}
Tibshirani, R., Saunders, M., Rosset, S., Zhu, J., Knight, K.: Sparsity and smoothness via the fused lasso. J. Roy. Stat. Soc. B (Stat.Methodol.), \textbf{67}(1), pp. 91--108 (2005)

\bibitem{AdaptiveLasso}
Zou, H.: The adaptive lasso and its oracle properties. J. Amer. Stat. Assoc., \textbf{101}(476), pp. 1418–-1429 (2006)

\bibitem{RelaxedLasso}
Meinshausen, N.: Relaxed lasso. Comput. Stat. Data Anal., \textbf{52}(1), pp. 374–-393 (2007)

\bibitem{HSICLasso}
Yamada, M.,  Jitkrittum, W., Sigal, L., Xing, E.P.,  Sugiyama, M.: High-dimensional feature selection by feature-wise kernelized lasso, Neural computation, \textbf{26}(1), pp. 185-–207, (2014)

\bibitem{NonParKer}
Liu, H., Wasserman, L., Lafferty, J. D.: Nonparametric regression and classification with joint sparsity constraints, Proc. Adv. Neural Inf. Process. Syst., pp. 969--976 (2009)

\bibitem{LLR}
Shevade S. K., Keerthi, S. S.: A simple and efficient algorithm for gene selection using sparse logistic regression, Bioinformatics, \textbf{19}, 17, pp. 2246--2253 (2003)

\bibitem{Latent1}
Chan, A.B., Vasconcelos, N., Lanckriet, G.R.G.: Direct convex relaxations
of sparse SVM. In: International Conference on Machine Learning, (2007)

\bibitem{Latent2}
Gurram, P., Kwon, H.: Optimal sparse kernel learning in the empirical kernel feature space for hyperspectral classification. IEEE Journal of Selected
Topics in Applied Earth Observations and Remote Sensing, (2014)


\bibitem{SCE} Ghosh, T., Karimov, K. ,Kirby, M.: Sparse Linear Centroid-Encoder: A Biomarker Selection tool for High Dimensional Biological Data. IEEE International Conference on Bioinformatics and Biomedicine (BIBM), pp. 3012--3019 (2023)


\bibitem{Hara}
O’Hara, S., Wang, K., Slayden, R.A. et all.: Iterative feature removal yields highly discriminative pathways. BMC genomics, \textbf{14}(1), pp. 1--15, (2013)

\bibitem{MTL}
Caruana, R.: Multitask learning, Machine learning, \textbf{28}(1), pp. 41--75 (1997) 

\bibitem{MDL}
Joshi, M., Cohen, W. W., Dredze M., Rosé, C.P.: Multi-domain learning: when do domains matter?, Proceedings of the 2012 Joint Conference on Empirical Methods in Natural Language Processing and Computational Natural Language Learning, pp. 1302--1312 (2012)

\bibitem{MDMT}
Yang, Y., Hospedales, T.M.: A unified perspective on multi-domain and multi-task learning. ArXiv, \textbf{1412.7489}, \url{https://arxiv.org/abs/1412.7489} (2014)

\bibitem{MTLDeep}
Ruder, S., An Overview of Multi-Task Learning in Deep Neural Networks. ArXiv, \textbf{1706.05098}, \url{https://arxiv.org/abs/1706.05098} (2017)

\bibitem{Uhler}
Yang, K.D., Belyaeva, A., Venkatachalapathy, S. et al.: Multi-domain translation between single-cell imaging and sequencing data using autoencoders. Nat Commun \textbf{12}(31) (2021). \doi{10.1038/s41467-020-20249-2}

\bibitem{VAE}
Kingma, D.P., Welling, M.: Auto-Encoding Variational Bayes. ArXiv, \textbf{1312.6114}, \url{https://arxiv.org/abs/1312.6114} (2013)

\bibitem{balance1}
Chen, Z., Badrinarayanan, V., Lee, C.Y., Rabinovich, A.: GradNorm: Gradient Normalization for Adaptive Loss Balancing in Deep Multitask Networks. International Conference on Machine Learning (2017)

\bibitem{balance2}
Sener, O., Koltun, V.: Multi-Task Learning as Multi-Objective Optimization. Neural Information Processing Systems (2018)

\bibitem{balance3}
Guo, M., Haque, A., Huang, DA., Yeung, S., Fei-Fei, L.: Dynamic Task Prioritization for Multitask Learning. Computer Vision – ECCV 2018, Lecture Notes in Computer Science, \textbf{11220}, Springer, Cham, \doi{10.1007/978-3-030-01270-0_17} (2018)




\bibitem{Code}
\url{https://github.com/kkarimov/iccs2024}

\bibitem{Ray}
\url{https://github.com/ray-project/ray}

\end{thebibliography}
\end{document}